\documentclass[11pt,a4paper]{article}
\usepackage[hyperref]{acl2019}
\usepackage{times}
\usepackage{latexsym}
\usepackage{tabularx}
\usepackage{graphicx}
\usepackage{float}
\usepackage{enumerate}
\usepackage{amssymb,amsmath}
\usepackage{color}
\usepackage{bm}
\usepackage{makecell}
\usepackage{multi row}
\usepackage{array}
\usepackage{url}
\usepackage{resizegather}
\usepackage{refcount}
\usepackage{fixfoot}

\newif\ifcomment
\commenttrue
\ifcomment
\newcommand{\kc}[1]{\textcolor{red}{KC: #1}}
\newcommand{\uk}[1]{\textcolor{blue}{UK: #1}}
\newcommand{\tl}[1]{\textcolor{green}{TL: #1}}
\newcommand{\ql}[1]{\textcolor{cyan}{QL: #1}}
\newcommand{\cm}[1]{\textcolor{orange}{CM: #1}}
\else
\newcommand{\kc}[1]{}
\newcommand{\uk}[1]{}
\newcommand{\tl}[1]{}
\newcommand{\ql}[1]{}
\newcommand{\cm}[1]{}
\fi

\newcommand\tstrut{\rule{0pt}{2.6ex}}         
\newcommand\bstrut{\rule[-1.0ex]{0pt}{0pt}}   
\newcommand{\thinline}{\Xhline{1.5\arrayrulewidth}}
\newcommand{\thickline}{\Xhline{2.5\arrayrulewidth}}
\newcommand{\tsep}	{\bstrut \\ \thinline}
\newcommand{\ttop}{\thickline}
\newcommand{\tbottom}{\bstrut \\ \thickline}

\newcommand{\xhdr}[1]{\vspace{1.7mm}\noindent{{\bf #1}}}

\newcommand{\alns}[1] {
	\begin{align*} #1 \end{align*}
}
\newcommand{\task}{\tau}
\newcommand{\tasks}{\mathcal{T}}
\newcommand{\dataset}{\mathcal{D}}
\newcommand{\loss}{\mathcal{L}}
\newcommand{\lss}{\ell}
\newcommand{\model}[1] {f_\task(#1)}
\newcommand{\modelabbrev}{f_\task}
\aclfinalcopy 
\def\aclpaperid{1484} 

\title{BAM! Born-Again Multi-Task Networks for Natural Language Understanding}

 \author{Kevin Clark$^\dagger$ \hspace{4mm} Minh-Thang Luong$^\ddagger$ \hspace{4mm} Urvashi Khandelwal$^\dagger$\\
 \textbf{Christopher D. Manning}$^\dagger$ \hspace{4mm} \textbf{Quoc V. Le}$^\ddagger$\\
  $^\dagger$Computer Science Department, Stanford University\\ $^\ddagger$Google Brain \\
  {\tt \{kevclark,urvashik,manning\}@cs.stanford.edu} \\ {\tt \{thangluong,qvl\}@google.com}
 }
\date{}

\begin{document}
\maketitle
\begin{abstract}
  It can be challenging to train multi-task neural networks that outperform or even match their single-task counterparts. 
  To help address this, we propose 
  using knowledge distillation where single-task models teach a multi-task model.
  We enhance this training with teacher annealing, a novel method that gradually transitions the model from distillation to supervised learning, helping the multi-task model surpass its single-task teachers.
  We evaluate our approach by multi-task fine-tuning BERT on the GLUE benchmark. 
  Our method consistently improves over standard single-task and multi-task training. 
\end{abstract}

\section{Introduction}
Building a single model that jointly learns to perform many tasks effectively has been a long-standing challenge in Natural Language Processing (NLP).
However, applying multi-task NLP remains difficult for many applications, with multi-task models often performing worse than their single-task counterparts \citep{Plank2017WhenIM, Bingel2017IdentifyingBT, mccann2018natural}. 
Motivated by these results, we propose a way of applying knowledge distillation \citep{buciluǎ2006model,Ba2014DoDN,hinton2015distilling} so that single-task models effectively teach a multi-task model.

Knowledge distillation transfers knowledge from a ``teacher" model to a ``student" model by training the student to imitate the teacher's outputs. 
In ``born-again networks" \citep{furlanello2018born}, the teacher and student have the same neural architecture and model size, but surprisingly the student is able to surpass the teacher's accuracy.
Intuitively, distillation is effective because the teacher's output distribution over classes provides more training signal than a one-hot label; \citet{hinton2015distilling} suggest that 
teacher outputs contain ``dark knowledge" capturing additional information about training examples.

\begin{figure}[t!]
\begin{center}
\includegraphics[width=0.999\linewidth]{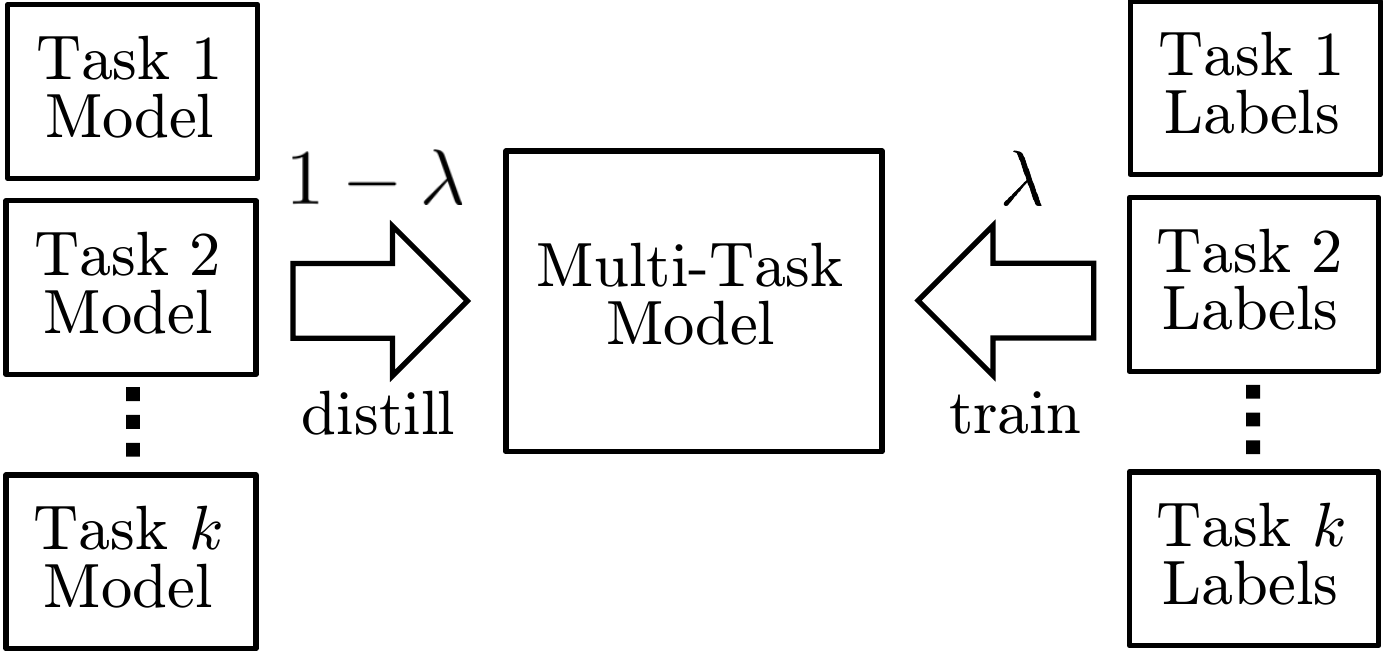}
\end{center}
\caption{An overview of our method. $\lambda$ is increased linearly from 0 to 1 over the course of training.}
\label{fig:model}
\end{figure}

Our work extends born-again networks to the multi-task setting.
We compare Single$\to$Multi\footnote{We use Single$\to$Multi to indicate distilling single-task ``teacher" models into a multi-task ``student" model.} born-again distillation with several other variants 
(Single$\to$Single and Multi$\to$Multi), and also explore performing multiple rounds of distillation (Single$\to$Multi$\to$Single$\to$Multi).
Furthermore, we propose a simple teacher annealing method that helps the student model outperform its teachers.
Teacher annealing gradually transitions the student from learning from the teacher to learning from the gold labels.
This method ensures the student gets a rich training signal early in training but is not limited to only imitating the teacher.

Our experiments build upon recent success in self-supervised pre-training \cite{dai2015semi,peters2018deep} and multi-task fine-tune BERT \cite{devlin2018bert} to perform the tasks from the GLUE natural language understanding benchmark \cite{wang2018glue}.
Our training method, which we call Born-Again Multi-tasking (BAM)\footnote{Code is available at \url{https://github.com/google-research/google-research/tree/master/bam}}, consistently outperforms standard single-task and multi-task training. 
Further analysis shows the multi-task models benefit from both better regularization and transfer between related tasks.

\section{Related Work}
Multi-task learning for neural networks in general \citep{Caruana1997MultitaskL} and within NLP specifically \citep{Collobert2008AUA,Luong2015MultitaskST} has been widely studied.
Much of the recent work for NLP has centered on neural architecture design: e.g., ensuring only beneficial information is shared across tasks \citep{Liu2017AdversarialML, Ruder2017SluiceNL}
or arranging tasks in linguistically-motivated hierarchies \citep{Sgaard2016DeepML, hashimoto2016joint, sanh2018hierarchical}.
These contributions are orthogonal to ours because we instead focus on the multi-task training algorithm.

Distilling large models into small models \citep{Kim2016SequenceLevelKD, mou2016distilling} or ensembles of models into single models \citep{Kuncoro2016DistillingAE,Liu2019ImprovingMD} has been shown to improve results for many NLP tasks.
There has also been some work on using knowledge distillation to aide in multi-task learning. 
In reinforcement learning, knowledge distillation has been used to regularize multi-task agents \citep{Parisotto2016ActorMimicDM, teh2017distral}. In NLP, \citet{tan2018multilingual} distill single-language-pair machine translation systems into a many-language system. However, they focus on multilingual rather than multi-task learning, use a more complex training procedure, and only experiment with Single$\to$Multi distillation.

Concurrently with our work, several other recent works also explore fine-tuning BERT using multiple tasks \citep{phang2018sentence,liu2019multi, keskar2019unifying}. However, they use only standard transfer or multi-task learning, instead focusing on finding beneficial task pairs or designing improved task-specific components on top of BERT.  

\section{Methods}
\subsection{Multi-Task Setup}
\xhdr{Model.} All of our models are built on top of BERT \citep{devlin2018bert}. This model passes byte-pair-tokenized \citep{Sennrich2016NeuralMT} input sentences through a Transformer network \citep{Vaswani2017AttentionIA}, producing a contextualized representation for each token. The vector corresponding to the first input token\footnote{For BERT this is a special token \texttt{[CLS]} that is prepended to each input sequence.}
$c$ is passed into a task-specific classifier. For classification tasks, we use a standard softmax layer: $\text{softmax}(W c)$. For regression tasks, we normalize the labels so they are between 0 and 1 and then use a size-1 NN layer with a sigmoid activation: $\text{sigmoid}(w^T c)$. 
In our multi-task models, all of the model parameters are shared across tasks except for these classifiers on top of BERT, which means less than 0.01\% of the parameters are task-specific.
Following BERT, the token embeddings and Transformer are initialized with weights from a self-supervised pre-training phase.

\xhdr{Training.} Single-task training is performed as in \citet{devlin2018bert}. For multi-task training, examples of different tasks are shuffled together, even within minibatches. The summed loss across all tasks is minimized.

\subsection{Knowledge Distillation}
We use $\dataset_\task = \{(x_\tau^1, y_\tau^1), ..., (x_\tau^N, y_\tau^N)\}$ to denote the training set for a task $\tau$ and $\model{x, \theta}$ to denote the outputs
for task $\tau$ produced by a neural network with parameters $\theta$ on the input $x$ (for classification tasks this is a distribution over classes). Standard supervised learning trains $\theta$ to minimize the loss on the training set:
\alns{
\loss(\theta) = \sum_{x^i_\tau, y^i_\tau \in \dataset_\tau} \lss(y^i_\tau, \model{x^i_\tau, \theta})
}
 where for classification tasks $\lss$ is usually cross-entropy. Knowledge distillation trains the model to instead match the predictions of a teacher model with parameters $\theta'$:
\alns{
\loss(\theta) = \sum_{x^i_\tau, y^i_\tau \in \dataset_\tau} \lss(\model{x^i_\tau, \theta'}, \model{x^i_\tau, \theta})
}
Note that our distilled networks are ``born-again" in that the student has the same model architecture as the teacher, i.e., all of our models have the same prediction function $\modelabbrev$ for each task.
For regression tasks, we train the student to minimize the L2 distance between its prediction and the teacher's instead of using cross-entropy loss. 
Intuitively, knowledge distillation improves training because the full distribution over labels provided by the teacher provides a richer training signal than a one-hot label. See \citet{furlanello2018born} for a more thorough discussion.

\xhdr{Multi-Task Distillation.}
Given a set of tasks $\tasks$, we train a single-task model with parameters $\theta_\tau$ on each task $\tau$. For most experiments, we use the single-task models to teach a multi-task model with parameters $\theta$:
\begin{gather*}
\loss(\theta) = \sum_{\task \in \tasks} \sum_{x^i_\tau, y^i_\tau \in \dataset_\tau} \lss(\model{x^i_\tau, \theta_\task}, \model{x^i_\tau, \theta})
\end{gather*}
However, we experiment with other distillation strategies as well. 

\addtolength{\tabcolsep}{-2pt}
\begin{table*}[t!]
\begin{center}
\begin{tabularx}{0.99\linewidth}{X >{\hsize=1.2cm}X l l l l l l l l}
\ttop 
\multirow{2}{*}{\textbf{Model}} & \multirow{2}{*}{\textbf{Avg.}} & \textbf{CoLA}$^\text{a}$ & \textbf{SST-2}$^\text{b}$ & \textbf{MRPC}$^\text{c}$ & \textbf{STS-B}$^\text{d}$ & \textbf{QQP}$^\text{e}$ & \textbf{MNLI}$^\text{f}$ & \textbf{QNLI}$^\text{g}$ & \textbf{RTE}$^\text{h}$ \tstrut \\ 
      &      & \hspace{-2mm}$|\dataset|$ = 8.5k & 67k & 3.7k & 5.8k & 364k & 393k & 108k & 2.5k\tsep
Single            & 84.0 & 60.6 & 93.2 & 88.0 & 90.0 & 91.3 & 86.6 & 92.3 & 70.4 \tstrut \\
Multi             & 85.5 & 60.3 & 93.3 & 88.0 & 89.8 & 91.4 & 86.5 & 92.2 & 82.1  \\
Single$\to$Single & 84.3 & $\textbf{61.7}^{**}$ & 93.2 & $\textbf{88.7}^{*}$ & 90.0 & 91.4 & $\textbf{86.8}^{**}$ & $\textbf{92.5}^{***}$ & 70.0 \\
Multi$\to$Multi   & 85.6 & 60.9 & 93.5 & 88.1 & 89.8 & $\textbf{91.5}^{*}$ & 86.7 & 92.3 & 82.0  \\
Single$\to$Multi  & $\textbf{86.0}^{***}$ & $\textbf{61.8}^{**}$ & $\textbf{93.6}^{*}$ & $\textbf{89.3}^{**}$ & 89.7 & $\textbf{91.6}^{*}$ & $\textbf{87.0}^{***}$ & $\textbf{92.5}^{***}$ & $\textbf{82.8}^{*}$ \tbottom
\end{tabularx} 
\end{center}
\small{Dataset references: $^\text{a}$\citet{Warstadt2018NeuralNA} $^\text{b}$\citet{Socher2013RecursiveDM} $^\text{c}$\citet{Dolan2005AutomaticallyCA} $^\text{d}$\citet{Cer2017SemEval2017T1} $^\text{e}$\citet{QQP} $^\text{f}$\citet{Williams2018ABC} $^\text{g}$constructed from SQuAD \citep{Rajpurkar2016SQuAD10}}
$^\text{h}$\citet{Giampiccolo2007TheTP}

\caption{Comparison of methods on the GLUE dev set. $^{*}$, $^{**}$, and $^{***}$ indicate statistically significant ($p < .05$, $p < .01$, and $p < .001$) improvements over both Single and Multi according to bootstrap hypothesis tests.\textsuperscript{\ref{hb}}}
\label{tab:main}
\end{table*}
\addtolength{\tabcolsep}{2pt}

\xhdr{Teacher Annealing.}
In knowledge distillation, the student is trained to imitate the teacher. 
This raises the concern that the student may be limited by the teacher's performance and not be able to substantially outperform the teacher.
To address this, we propose {\it teacher annealing}, which mixes the teacher prediction with the gold label during training. Specifically, the term in the summation becomes
\begin{align*}
\lss(\lambda y^i_\tau + (1 - \lambda) \model{x^i_\tau, \theta_\task}, \model{x^i_\tau, \theta})
\end{align*}

\noindent where $\lambda$ is linearly increased from 0 to 1 throughout training. Early in training, the model is mostly distilling to get as useful of a training signal as possible. Towards the end of training, the model is mostly relying on the gold-standard labels so it can learn to surpass its teachers.

\section{Experiments}

\xhdr{Data.} We use the General Language Understanding Evaluation (GLUE) benchmark \citep{wang2018glue}, which consists of 9 natural language understanding tasks on English data.
Tasks cover textual entailment (RTE and MNLI) question-answer entailment (QNLI), paraphrase (MRPC), question paraphrase (QQP), textual similarity (STS), sentiment (SST-2), linguistic acceptability (CoLA), and Winograd Schema (WNLI).

\xhdr{Training Details.} 
Rather than simply shuffling the datasets for our multi-task models, we follow the task sampling procedure from \citet{bowman2018looking}, where the probability of training on an example for a particular task $\task$ is proportional to $|\dataset_\task|^{0.75}$. This ensures that tasks with very large datasets don't overly dominate the training. 

We also use the layerwise-learning-rate trick from \citet{howard2018universal}. If layer 0 is the NN layer closest to the output, the learning rate for a particular layer $d$ is set to $\textsc{base\_lr} \cdot \alpha^d$ (i.e., layers closest to the input get lower learning rates). The intuition is that pre-trained layers closer to the input learn more general features, so they shouldn't be altered much during training. 

\xhdr{Hyperparameters.} For single-task models, we use the same hyperparameters as in the original BERT experiments
except we pick a layerwise-learning-rate decay $\alpha$ of 1.0 or 0.9 on the dev set for each task. 
For multi-task models, we train the model for longer (6 epochs instead of 3) and with a larger batch size (128 instead of 32), using $\alpha=0.9$ and a learning rate of 1e-4. 
All models use the BERT-Large pre-trained weights.

\xhdr{Reporting Results.}
Dev set results report the average score (Spearman correlation for STS, Matthews correlation for CoLA, and accuracy for the other tasks) on all GLUE tasks except WNLI, for which methods can't outperform a majority baseline. Results show the median score of at least 20 trials with different random seeds. 
We find using a large number of trials is essential because results can vary significantly for different runs. For example, standard deviations in score are over $\pm1$ for CoLA, RTE, and MRPC for multi-task models. Single-task standard deviations are even larger.

\section{Results}
\xhdr{Main Results.} 
We compare models trained with single-task learning, multi-task learning, and several varieties of distillation in Table~\ref{tab:main}. While standard multi-task training improves over single-task training for RTE (likely because it is closely related to MNLI), there is no improvement on the other tasks. 
In contrast, Single$\to$Multi knowledge distillation improves or matches the performance of the other methods on all tasks except STS, the only regression task in GLUE. 
We believe distillation does not work well for regression tasks because there is no distribution over classes passed on by the teacher to aid learning. 

The gain for Single$\to$Multi over Multi is larger than the gain for Single$\to$Single over Single, suggesting that distillation works particularly well in combination with multi-task learning.
Interestingly, Single$\to$Multi works substantially better than Multi$\to$Multi distillation.
We speculate it may help that the student is exposed to a diverse set of teachers in the same way ensembles benefit from a diverse set of models, but future work is required to fully understand this phenomenon. 
In addition to the models reported in the table, we also trained Single$\to$Multi$\to$Single$\to$Multi models. However, the difference with Single$\to$Multi was not statistically significant, suggesting there is little value in multiple rounds of distillation.

Overall, a key benefit of our method is robustness: while standard multi-task learning produces mixed results, Single$\to$Multi distillation  consistently outperforms standard single-task and multi-task training.
We also note that in some trials single-task training resulted in models that score quite poorly (e.g., less than 91 for QQP or less than 70 for MRPC), while the multi-task models have more dependable performance.

\footnotetext{\label{hb}For all statistical tests we use the Holm-Bonferroni method \citep{holm1979simple} to correct for multiple comparisons.}

\addtolength{\tabcolsep}{-8pt}
\begin{table}
\begin{tabularx}{\linewidth}{ l c }
\ttop
\textbf{Model} & \textbf{GLUE score} \tstrut \tsep
BERT-Base \cite{devlin2018bert} & 78.5\tstrut \\
BERT-Large \cite{devlin2018bert} & 80.5 \\
BERT on STILTs \cite{phang2018sentence} & 82.0 \\
MT-DNN  \cite{liu2019multi}  & 82.2\\
Span-Extractive BERT on STILTs \vspace{-0.8mm} & \multirow{2}{*}{82.3} \\ \cite{keskar2019unifying} &  \\
Snorkel MeTaL ensemble \vspace{-0.8mm} & \multirow{2}{*}{83.2} \\
\cite{Snorkel} & \\
MT-DNN$_{KD}$* \cite{Liu2019ImprovingMD} & 83.7\tsep
BERT-Large + BAM \textbf{(ours)} & 82.3\tstrut
\tbottom
\end{tabularx}
\caption{Comparison of test set results. *MT-DNN$_{KD}$ is distilled from a diverse ensemble of models.}
\label{tab:test}
\end{table}
\addtolength{\tabcolsep}{8pt}

\xhdr{Test Set Results.}
We compare against recent work by submitting to the GLUE leaderboard. 
We use Single$\to$Multi distillation.
Following the procedure used by BERT, we train multiple models and submit the one with the highest average dev set score to the test set. BERT trained 10 models for each task (80 total); we trained 20 multi-task models. Results are shown in Table~\ref{tab:test}.

Our work outperforms or matches existing published results that do not rely on ensembling.
However, due to the variance between trials discussed under ``Reporting Results," we think these test set numbers should be taken with a grain of salt, as they only show the performance of individual training runs (which is further complicated by the use of tricks such as dev set model selection). We believe significance testing over multiple trials would be needed to have a definitive comparison.

\xhdr{Single-Task Fine-Tuning}.
A crucial difference distinguishing our work from the STILTs, Snorkel MeTaL, and MT-DNN$_{KD}$ methods in Table~\ref{tab:test} is that we do not single-task fine-tune our model. 
That is, we do not further train the model on individual tasks after the multi-task training finishes. 
While single-task fine-tuning improves results, we think to some extent it defeats the purpose of multi-task learning: the result of training is one model for each task instead of a model that can  perform all of the tasks.
Compared to having many single-task models, a multi-task model is simpler to deploy, faster to run, and arguably more scientifically interesting from the perspective of building general language-processing systems.

We evaluate the benefits of single-task fine-tuning in Table~\ref{tab:finetune}. 
Single-task fine-tuning initializes models with multi-task-learned weights and then performs single-task training. Hyperparameters are the same as for our single-task models except we use a smaller learning rate of 1e-5. 
While single-task fine-tuning unsurprisingly improves results, the gain on top of Single$\to$Multi distillation is small, reinforcing the claim that distillation provides many of the benefits of single-task training while producing a single unified model instead of many task-specific models.

\newcolumntype{Y}{>{\centering\arraybackslash}X}
\begin{table}
\begin{tabularx}{\linewidth}{X >{\hsize=1.85cm}Y}
\ttop 
\textbf{Model} & \textbf{Avg. Score} \tstrut \tsep
Multi & 85.5 \tstrut  \\
\hspace{2mm} +Single-Task Fine-Tuning & $+$0.3 \\
Single$\to$Multi & 86.0  \\
\hspace{2mm}  +Single-Task Fine-Tuning & $+$0.1\tbottom
\end{tabularx}
\caption{Combining multi-task training with single-task fine-tuning. Improvements are statistically significant ($p < .01$) according to Mann-Whitney U tests.\textsuperscript{\ref{hb}}
}
\label{tab:finetune}
\end{table}

\xhdr{Ablation Study.}
We show the importance of teacher annealing and the other training tricks in Table~\ref{tab:ablation}. We found them all to significantly improve scores.
Using pure distillation without teacher annealing (i.e., fixing $\lambda = 0$) performs no better than standard multi-task learning, demonstrating the importance of the proposed teacher annealing method.

\begin{table}
\begin{tabularx}{\linewidth}{X >{\hsize=1.85cm}Y}
\ttop 
\textbf{Model} & \textbf{Avg. Score} \tstrut \tsep
Single$\to$Multi & 86.0 \tstrut  \\
\hspace{2mm} No layer-wise LRs & $-$0.3 \\
\hspace{2mm} No task sampling & $-$0.4 \\
\hspace{2mm} No teacher annealing: $\lambda = 0$ & $-$0.5 \\ 
\hspace{2mm} No teacher annealing: $\lambda = 0.5$ & $-$0.3\tbottom
\end{tabularx}
\caption{Ablation Study. Differences from Single$\to$Multi are statistically significant ($p < .001$) according to Mann-Whitney U tests.\textsuperscript{\ref{hb}}
}
\label{tab:ablation}
\end{table}

\xhdr{Comparing Combinations of Tasks.}
Training on a large number of tasks is known to help regularize multi-task models \citep{Ruder2017AnOO}. A related benefit of multi-task learning is the transfer of learned ``knowledge" between closely related tasks.
We investigate these two benefits by comparing several models on the RTE task, including one trained with a very closely related task (MNLI, a much large textual entailment dataset) and one trained with fairly unrelated tasks (QQP, CoLA, and SST). 
We use Single$\to$Multi distillation (Single$\to$Single in the case of the RTE-only model).
Results are shown in Table~\ref{tab:reg}.
We find both sets of auxiliary tasks improve RTE performance, suggesting that both benefits are playing a role in improving multi-task models.
Interestingly, RTE + MNLI alone slightly outperforms the model performing all tasks, perhaps because training on MNLI, which has a very large dataset, is already enough to sufficiently regularize the model. 

\begin{table}
\begin{tabularx}{\linewidth}{X >{\hsize=2cm}Y}
\ttop 
\textbf{Trained Tasks} & \textbf{RTE score} \tstrut \tsep
RTE & 70.0\tstrut \\
RTE + MNLI & 83.4 \\
RTE + QQP + CoLA + SST & 75.1 \\
All GLUE & 82.8\tbottom
\end{tabularx}
\caption{Which tasks help RTE? Pairwise differences are statistically significant $(p < .01$) according to Mann-Whitney U tests.\textsuperscript{\ref{hb}}}
\label{tab:reg}
\end{table}

\section{Discussion and Conclusion}
We have shown that  Single$\to$Multi distillation combined with teacher annealing produces results consistently better than standard single-task or multi-task training.
Achieving robust multi-task gains across many tasks has remained elusive in previous research, so we hope our work will make multi-task learning more broadly useful within NLP\@.
However, with the exception of closely related tasks with small datasets (e.g., MNLI helping RTE), the overall size of the gains from our multi-task method are small compared to the gains provided by transfer learning from self-supervised tasks (i.e., BERT).
It remains to be fully understood to what extent ``self-supervised pre-training is all you need" and where transfer/multi-task learning from supervised tasks can provide the most value.

\section*{Acknowledgements}
We thank Robin Jia, John Hewitt, and the
anonymous reviewers for their thoughtful comments and suggestions.
Kevin is supported by a
Google PhD Fellowship.

\bibliography{bam}
\bibliographystyle{acl_natbib}

\end{document}